\documentclass[11pt, a4paper]{article}
\usepackage[table]{xcolor}
\usepackage[hyperref]{emnlp2018}
\usepackage{CJKutf8}
\usepackage{times}
\usepackage{latexsym}
\usepackage{tikz}
\usetikzlibrary{shapes,arrows,chains}
\usepackage{tikz-dependency}
\usepackage{url}
\usepackage{multirow}
\usepackage{array}
\usepackage{subcaption}
\usepackage{float}
\usepackage{setspace}
\usepackage{verbatim}
\usepackage{lingmacros}
\usepackage{forest}
\useforestlibrary{linguistics}
\forestapplylibrarydefaults{linguistics}
\usepackage{linguex}
\usepackage{booktabs}

\forestset{
sm edges/.style={for tree={parent anchor=south, child anchor=north,base=bottom},
                 where n children=1{tier=word}{}
                 },
}   

\forestset{default preamble'={
    for tree={align=center,parent anchor=south, child anchor=north,base=bottom},
    before drawing tree={
      sort by=y,
      for min={tree}{baseline}
    }
}} 

\aclfinalcopy 

\title{Semantic Role Labeling for Learner Chinese: \\the Importance of Syntactic Parsing and L2-L1 Parallel Data}

\author{Zi Lin$^{\text{123}}$, Yuguang Duan$^{\text{3}}$, Yuanyuan Zhao$^{\text{125}}$, Weiwei Sun$^{\text{124}}$ and Xiaojun Wan$^{\text{12}}$\\
$^{\text{1}}$Institute of Computer Science and Technology, Peking University \\
$^{\text{2}}$The MOE Key Laboratory of Computational Linguistics, Peking University \\
$^{\text{3}}$Department of Chinese Language and Literature, Peking University\\
$^{\text{4}}$Center for Chinese Linguistics, Peking University\\
$^{\text{5}}$Academy for Advanced Interdisciplinary Studies, Peking University\\
\texttt{\{\href{mailto:zi.lin@pku.edu.cn}{zi.lin},\href{mailto:ariaduan@pku.edu.cn}{ariaduan},\href{mailto:wa@pku.edu.cn}{ws},\href{mailto:wanxiaojun@pku.edu.cn}{wanxiaojun}\}@pku.edu.cn, \href{mailto:zhaoyy1461@gmail.com}{zhaoyy1461@gmail.com}}\\}

\date{}
\begin{document}

\maketitle
\begin{abstract}
This paper studies semantic parsing for interlanguage (L2\footnote{
  In this paper, we call sentences written by non-native speakers (henceforth, ``L2 sentences''), aligned to their corrections by native speakers (henceforth, ``L1 sentences'') L2-L1 parallel sentences.}),
taking semantic role labeling (SRL) as a case task and learner Chinese as a case language. 
  We first manually annotate the semantic roles for a set of learner texts to derive a gold standard for automatic SRL. 
  Based on the new data, we then evaluate three off-the-shelf SRL systems, i.e., the PCFGLA-parser-based, neural-parser-based and neural-syntax-agnostic systems, to gauge how successful SRL for learner Chinese can be.
  We find two non-obvious facts: 1) the L1-sentence-trained systems performs rather badly on the L2 data; 
  2) the performance drop from the L1 data to the L2 data of the two parser-based systems is much smaller, 
  indicating the importance of syntactic parsing in SRL for interlanguages.
  Finally, the paper introduces a new agreement-based model to explore the semantic coherency information in the large-scale L2-L1 parallel data.
  We then show such information is very effective to enhance SRL for learner texts.
  Our model achieves an F-score of 72.06, which is a 2.02 point improvement over the best baseline.

\end{abstract}
\begin{CJK*}{UTF8}{gbsn}
\section{Introduction}
\label{sec:intro}
A learner language (interlanguage) is an idiolect developed by a learner of a second or foreign language which may preserve some features of his/her first language. 
Previously, encouraging results of automatically building the syntactic analysis of learner languages were reported \cite{nagata-sakaguchi:2016:P16-1}, 
but it is still unknown how semantic processing performs, while parsing a learner language (L2) into semantic representations is the foundation of a variety of deeper analysis of learner languages, e.g., automatic essay scoring.
In this paper, we study semantic parsing for interlanguage, taking semantic role labeling (SRL) as a case task and learner Chinese as a case language.

Before discussing a computation system, we first consider the linguistic competence and performance.
Can human robustly understand learner texts? Or to be more precise, to what extent, a native speaker can understand the meaning of a sentence written by a language learner? 
Intuitively, the answer is towards the positive side.
To validate this, we ask two senior students majoring in Applied Linguistics to carefully annotate some L2-L1 parallel sentences with predicate--argument structures according to the specification of Chinese PropBank \citep[CPB;][]{ChPropBank}, which is developed for L1. 
A high inter-annotator agreement is achieved, suggesting the robustness of language comprehension for L2.
During the course of semantic annotation, we find a non-obvious fact that we can re-use the semantic annotation specification, Chinese PropBank in our case, which is developed for L1. 
Only modest rules are needed to handle some tricky phenomena. 
This is quite different from syntactic treebanking for learner sentences, where defining a rich set of new annotation heuristics seems necessary \cite{ragheb:dickinson:12,nagata-sakaguchi:2016:P16-1,berzak-EtAl:2016:P16-1}.

Our second concern is to mimic the human's robust semantic processing ability by computer programs. The feasibility of reusing the annotation specification for L1 implies that we can reuse {\it standard} CPB data to train an SRL system to process learner texts. 
To test the {\it robustness} of the state-of-the-art SRL algorithms, we evaluate two types of SRL frameworks. 
The first one is a traditional SRL system that leverages a syntactic parser and heavy feature engineering to obtain {\it explicit} information of semantic roles \citep{wsun:smt12}. 
Furthermore, we employ two different parsers for comparison: 1) the PCFGLA-based parser, viz. Berkeley parser \cite{Petrov:06}, and 2) a minimal span-based neural parser \cite{stern2017minimal}. 
The other SRL system uses a stacked BiLSTM to implicitly capture local and non-local information \citep{he2017deep}. and we call it the neural syntax-agnostic system. 
All systems can achieve state-of-the-art performance on L1 texts but show a significant degradation on L2 texts. 
This highlights the weakness of applying an L1-sentence-trained system to process learner texts.

While the neural syntax-agnostic system obtains superior performance on the L1 data, the two syntax-based systems both produce better analyses on the L2 data.
Furthermore, as illustrated in the comparison between different parsers, the better the parsing results we get, the better the performance on L2 we achieve. 
This shows that syntactic parsing is important in semantic construction for learner Chinese. 
The main reason, according to our analysis, is that the syntax-based system may generate correct syntactic analyses for partial grammatical fragments in L2 texts, which provides crucial information for SRL. 
Therefore, syntactic parsing helps build more generalizable SRL models that transfer better to new {\it languages}, and enhancing syntactic parsing can improve SRL to some extent.

Our last concern is to explore the potential of a large-scale set of L2-L1 parallel sentences to enhance SRL systems. 
We find that semantic structures of the L2-L1 parallel sentences are highly consistent. 
This inspires us to design a novel agreement-based model to explore such semantic coherency information. 
In particular, we define a metric for comparing predicate--argument structures and searching for relatively good automatic syntactic and semantic annotations to extend the training data for SRL systems.
Experiments demonstrate the value of the L2-L1 parallel sentences as well as the effectiveness of our method. 
We achieve an F-score of 72.06, which is a 2.02 percentage point improvement over the best neural-parser-based baseline. 

To the best of our knowledge, this is the first time that the L2-L1 parallel data is utilized to enhance NLP systems for learner texts. 

For research purpose, we have released our SRL annotations on 600 sentence pairs and the L2-L1 parallel dataset \footnote{The data is collected from Lang-8 (\url{www.lang-8.com}) and used as the training data in NLPCC 2018 Shared Task: Grammatical Error Correction \citep{nlpcc2018}, which can be downloaded at \url{https://github.com/pkucoli/srl4il}}.

\section{Semantic Analysis of An L2-L1 Parallel Corpus}
\label{sec:human}

\subsection{An L2-L1 Parallel Corpus}
An L2-L1 parallel corpus can greatly facilitate the analysis of a learner language \cite{l1l2treebank}. 
Following \newcite{mizumoto:2011}, we collected a large dataset of L2-L1 parallel texts of Mandarin Chinese by exploring ``language exchange" social networking services (SNS), 
i.e., Lang-8, a language-learning website where native speakers can freely correct the sentences written by foreign learners. The proficiency levels of the learners are diverse, but most of the learners, according to our judgment, is of intermediate or lower level.

Our initial collection consists of 1,108,907 sentence pairs from 135,754 essays. As there is lots of noise in raw sentences, we clean up the data by (1) ruling out redundant content, (2) excluding sentences containing foreign words or Chinese phonetic alphabet by checking the Unicode values, (3) dropping overly simple sentences which may not be informative, and (4) utilizing a rule-based classifier to determine whether to include the sentence into the corpus.

The final corpus consists of 717,241 learner sentences from writers of 61 different native languages, in which English and Japanese constitute the majority. As for completeness, 82.78\% of the Chinese Second Language sentences on Lang-8 are corrected by native human annotators. One sentence gets corrected approximately 1.53 times on average.

In this paper, we manually annotate the predicate--argument structures for the 600 L2-L1 pairs as the basis for the semantic analysis of learner Chinese. It is from the above corpus that we carefully select 600 pairs of L2-L1 parallel sentences. We would choose the most appropriate one among multiple versions of corrections and recorrect the L1s if necessary.
Because word structure is very fundamental for various NLP tasks, our annotation also contains {\it gold} word segmentation for both L2 and L1 sentences. 
Note that there are no natural word boundaries in Chinese text. We first employ a state-of-the-art word segmentation system to produce initial segmentation results and then manually fix segmentation errors.

The dataset includes four typologically different mother tongues, 
i.e., English (ENG), Japanese (JPN), Russian (RUS) and Arabic (ARA). 
Sub-corpus of each language consists of 150 sentence pairs.
We take the mother languages of the learners into consideration, which have a great impact on grammatical errors and hence automatic semantic analysis. 
We hope that four selected mother tongues guarantee a good coverage of typologies. 
The annotated corpus can be used both for linguistic investigation and as test data for NLP systems.

\subsection{The Annotation Process}
\begin{table}[]
\centering
\scalebox{0.95}{
\begin{tabular}{llccc}
\hline
        &&     P    &     R   &     F   \\\hline
\multirow{2}{*}{ENG}&  L1     &   95.87 &  96.17 &  96.02\\
&L2     &   94.78 &  93.06 &  93.91\\\hline
\multirow{2}{*}{JPN} &L1     &   97.95 &  98.69 &  98.32\\
&L2     &   96.07 &  97.48 &  96.77\\\hline
\multirow{2}{*}{RUS}&L1     &   96.95 &  95.41 &  96.17\\
 & L2     &   97.04 &  94.08 &  95.53\\\hline
\multirow{2}{*}{ARA}&L1     &   96.95 &  97.76 &  97.35\\
 & L2     &   97.12 &  97.56 &  97.34\\
\hline                       
\end{tabular}}
\caption{Inter-annotator agreement.}
\label{tb:agreement}
\end{table}

Semantic role labeling (SRL) is the process of assigning semantic roles to constituents or their head words in a sentence according to their relationship to the predicates expressed in the sentence.
Typical semantic roles can be divided into core arguments and adjuncts. 
The core arguments include {\it Agent}, {\it Patient}, {\it Source}, {\it Goal}, etc, while the adjuncts include {\it Location}, {\it Time}, {\it Manner}, {\it Cause}, etc.

To create a {\it standard} semantic-role-labeled corpus for learner Chinese, 
we first annotate a 50-sentence trial set for each native language. 
Two senior students majoring in Applied Linguistics conducted the annotation. 
Based on a total of 400 sentences, we adjudicate an initial gold standard, adapting and refining CPB specification as our annotation heuristics. 
Then the two annotators proceed to annotate a 100-sentence set for each language independently. 
It is on these larger sets that we report the inter-annotator agreement.

In the final stage, we also produce an adjudicated gold standard for all 600 annotated sentences. This was achieved by comparing the annotations selected by each annotator, discussing the differences, and either selecting one as fully correct or creating a hybrid representing the consensus decision for each choice point. When we felt that the decisions were not already fully guided by the existing annotation guidelines, we worked to articulate an extension to the guidelines that would support the decision.

During the annotation, the annotators apply both position labels and semantic role labels. 
Position labels include {\it S}, {\it B}, {\it I} and {\it E}, which are used to mark whether the word is an argument by itself, or at the beginning or in the middle or at the end of a argument.
As for role labels, we mainly apply representations defined by CPB \citep{ChPropBank}. 
The predicate in a sentence was labeled as \textit{rel}, the core semantic roles were labeled as \textit{AN} and the adjuncts were labeled as \textit{AM}.

\subsection{Inter-annotator Agreement}

For inter-annotator agreement, we evaluate the precision (P), recall (R), and F1-score (F) of the semantic labels given by the two annotators. Table \ref{tb:agreement} shows that our inter-annotator agreement is promising. All L1 texts have F-score above 95, and we take this as a reflection that our annotators are qualified. F-scores on L2 sentences are all above 90, just a little bit lower than those of L1, indicating that L2 sentences can be greatly understood by native speakers. Only modest rules are needed to handle some tricky phenomena:

\begin{enumerate}
\item The labeled argument should be strictly limited to the core roles defined in the frameset of CPB, though the number of arguments in L2 sentences may be more or less than the number defined.
\item For the roles in L2 that cannot be labeled as arguments under the specification of CPB, if they provide semantic information such as time, location and reason, we would labeled them as adjuncts though they may not be well-formed adjuncts due to the absence of function words.
\item For unnecessary roles in L2 caused by mistakes of verb subcategorization (see examples in Figure \ref{fig:output example2}), we would leave those roles unlabeled.
\end{enumerate}

Table \ref{tb:agreement-detail} further reports agreements on each argument (\textit{AN}) and adjunct (\textit{AM}) in detail, according to which the high scores are attributed to the high agreement on arguments (\textit{AN}). 
The labels of \textit{A3} and \textit{A4} have no disagreement since they are sparse in CPB and are usually used to label specific semantic roles that have little ambiguity. 

We also conducted in-depth analysis on inter-annotator disagreement. For further details, please refer to \newcite{duan2018argument}.

\begin{table}[H]
\centering
\scalebox{0.8}{
\begin{tabular}{llcccc}
\hline
                    &    & ENG    & JPN    & RUS    & ARA    \\ \hline
\multirow{6}{*}{L1} & \textit{A0} & 97.23  & 99.10  & 97.66  & 98.22  \\
                    & \textit{A1} & 96.70  & 96.99  & 98.05  & 98.34  \\
                    & \textit{A2} & 88.89  & 100.00 & 100.00 & 92.59  \\
                    & \textit{A3} & 100.00 & 100.00 & 100.00 & 100.00 \\
                    & \textit{A4} & 100.00 & -      & -      & 100.00 \\
                    & \textit{AM} & 94.94  & 98.35  & 93.07  & 96.02  \\ \hline
\multirow{6}{*}{L2} & \textit{A0} & 94.09  & 95.77  & 97.92  & 97.88  \\
                    & \textit{A1} & 90.68  & 97.93  & 97.40  & 98.68  \\
                    & \textit{A2} & 88.46  & 100.00 & 95.24  & 93.33  \\
                    & \textit{A3} & 100.00 & 100.00 & 100.00 & -      \\
                    & \textit{A4} & 100.00 & -      & -      & -      \\
                    & \textit{AM} & 96.97  & 96.51  & 91.78  & 96.02  \\ 
\hline
\end{tabular}}
\caption{Inter-annotator agreement (F-scores) relative to languages and role types.}
\label{tb:agreement-detail}
\end{table}

\section{Evaluating Robustness of SRL}
\begin{table*}[]
\centering
\scalebox{0.83}{
\begin{tabular}{llllllllllllllll}
\hline
 &  & \multicolumn{4}{c}{PCFGLA-parser-based SRL} & & \multicolumn{4}{c}{Neural-parser-based SRL}  & & \multicolumn{4}{c}{Neural syntax-agnostic SRL}\\
 \cline{3-6} \cline{8-11} \cline{13-16}
 &  & \multicolumn{1}{c}{Arg.-F} & \multicolumn{1}{c}{Adj.-F} & \multicolumn{1}{c}{F} & \multicolumn{1}{c}{$\Delta$F} & &\multicolumn{1}{c}{Arg.-F} & \multicolumn{1}{c}{Adj.-F} & \multicolumn{1}{c}{F} & \multicolumn{1}{c}{$\Delta$F} & &\multicolumn{1}{c}{Arg.-F} & \multicolumn{1}{c}{Adj.-F} & \multicolumn{1}{c}{F} & \multicolumn{1}{c}{$\Delta$F} \\\hline
\multirow{2}{*}{ENG} & L1 & 
73.42 & 74.60 & 73.81 & \multirow{2}{*}{-4.61} & & 
72.54 & 76.97& 74.00 & \multirow{2}{*}{-4.90} & & 
74.62 & 73.44 & 74.22 & \multirow{2}{*}{-6.65}\\
 & L2 & 
 68.05 & 71.63 & 69.20 & && 
 67.12 & 73.24 & 69.10 & && 
 66.50 & 69.75 & 67.57 \\\hline
 
\multirow{2}{*}{JPN} & L1 & 
73.88 & 76.80 & 75.10 & \multirow{2}{*}{-3.37} & & 
75.18 & 77.25& 76.05  & \multirow{2}{*}{-3.79} & &
76.50 & 78.22 & 77.22 & \multirow{2}{*}{-7.60}\\
 & L2 & 
 67.92 & 77.29 & 71.73 & && 
 69.59& 76.14& 72.26 & && 
 66.75 & 73.78 & 69.62  \\\hline
 
\multirow{2}{*}{RUS} & L1 & 
71.56 & 73.41 & 72.20 & \multirow{2}{*}{-4.28} & & 
73.20 & 73.72& 73.39  & \multirow{2}{*}{-3.01} & &
72.01 & 73.72 & 72.61 & \multirow{2}{*}{-5.60}\\
 & L2 & 
 68.41 & 66.93 & 67.92 &  & & 
 68.93 & 73.11& 70.38 & & &
 65.50 & 69.91 & 67.01 \\\hline

\multirow{2}{*}{ARA} & L1 & 
73.72 & 64.36 & 70.82 & \multirow{2}{*}{-2.18} & & 
75.38 &67.10 & 72.74  & \multirow{2}{*}{-2.80} & &
74.10 & 70.44 & 72.92 & \multirow{2}{*}{-2.40} \\
 & L2 & 
 69.43 & 67.02 & 68.64 & && 
 71.16 & 67.25& 69.94 & &&
 72.13 & 67.19 & 70.52 \\\hline

\multirow{2}{*}{ALL} & L1 & 
73.18 & 72.28 & 72.87 & \multirow{2}{*}{-3.59} & & 
74.05 & 73.73& 73.94  & \multirow{2}{*}{-3.64} & &
74.22 & 73.92 & 74.12 & \multirow{2}{*}{-5.41}  \\
 & L2 & 
 68.52 & 70.77 & 69.28 & && 
 69.20& 72.39  & 70.30 & & & 
 67.99 & 70.08 & 68.71 \\\hline
\end{tabular}}
\caption{Performances of the syntax-based and neural syntax-agnostic SRL systems on the L1 and L2 data. ``ALL'' denotes the overall performance.}
\label{overall system accuracy}
\end{table*}
\subsection{Three SRL Systems}
The work on SRL has included a broad spectrum of machine learning and deep learning approaches to the task. Early work showed that syntactic information is crucial for learning long-range dependencies, syntactic constituency structure and global constraints \cite{Punyakanok:08,tackstrom2015efficient}, 
while initial studies on neural methods achieved state-of-the-art results with little to no syntactic input \cite{zhou2015end,wang2015chinese,marcheggiani2017simple,he2017deep}. 
However, the question whether fully labeled syntactic structures provide an improvement for neural SRL is still unsettled pending further investigation.

To evaluate the {\it robustness} of state-of-the-art SRL algorithms, we evaluate two representative SRL frameworks.
One is a traditional syntax-based SRL system that leverages a syntactic parser and manually crafted features to obtain {\it explicit} information to find semantic roles \citep{Gildea:00,Xue:08}
In particular, we employ the system introduced in \citet{wsun:smt12}. 
This system first collects all c-commanders of a predicate in question from the output of a parser and puts them in order. 
It then employs a first order linear-chain global linear model to perform semantic tagging.
For constituent parsing, we use two parsers for comparison, one is Berkeley parser\footnote{\url{code.google.com/p/berkeleyparser/}} \citep{Petrov:06}, 
a well-known implementation of the unlexicalized latent variable PCFG model,
the other is a minimal span-based neural parser based on independent scoring of labels and spans \cite{stern2017minimal}. 
As proposed in \citet{stern2017minimal}, the second parser is capable of achieving state-of-the-art single-model performance on the Penn Treebank.
On the Chinese TreeBank \citep[CTB;][]{ctb}, it also outperforms the Berkeley parser for the in-domain test.
We call the corresponding SRL systems as the {\bf PCFGLA-parser-based} and {\bf neural-parser-based} systems. 

The second SRL framework leverages an end-to-end neural model to implicitly capture local and non-local information \citep{zhou2015end,he2017deep}.
In particular, this framework treats SRL as a {\it BIO} tagging problem and uses a stacked BiLSTM to find informative embeddings.
We apply the system introduced in \citet{he2017deep} for experiments. 
Because all syntactic information (including POS tags) is excluded, we call this system the {\bf neural syntax-agnostic} system.

To train the three SRL systems as well as the supporting parsers, we use the CTB and CPB data \footnote{Here we only use the trees that has semantic role annotations for parser training. This setup keeps us from overestimating the contribution of a parser.}.
In particular, the sentences selected for the CoNLL 2009 shared task are used here for parameter estimation. 
Note that, since the Berkeley parser is based on PCFGLA grammar, it may fail to get the syntactic outputs for some sentences, 
while the other parser does not have that problem. In this case, we have made sure that both parsers can parse all 1,200 sentences successfully.

\subsection{Main Results}
The overall performances of the three SRL systems on both L1 and L2 data (150 parallel sentences for each mother tongue) are shown in Table \ref{overall system accuracy}.
For all systems, significant decreases on different mother languages can be consistently observed, 
highlighting the weakness of applying L1-sentence-trained systems to process learner texts. 
Comparing the two syntax-based systems with the neural syntax-agnostic system, we find that the overall $\Delta$F, 
which denotes the F-score drop from L1 to L2, is smaller in the syntax-based framework than in the syntax-agnostic system.
On English, Japanese and Russian L2 sentences, the syntax-based system has better performances though it sometimes works worse on the corresponding L1 sentences, 
indicating the syntax-based systems are more robust when handling learner texts.

Furthermore, the neural-parser-based system achieves the best overall performance on the L2 data.
Though performing slightly worse than the neural syntax-agnostic one on the L1 data, it has much smaller $\Delta$F, 
showing that as the syntactic analysis improves, the performances on both the L1 and L2 data grow, while the gap can be maintained. 
This demonstrates again the importance of syntax in semantic constructions, especially for learner texts.

\subsection{Analysis}
To better understand the overall results, we further look deep into the output by addressing the questions: 

\begin{enumerate}
\item What types of error negatively impact both systems over learner texts? 
\item What types of error are more problematic for the neural syntax-agnostic one over the L2 data but can be solved by the syntax-based one to some extent?
\end{enumerate}

We first carry out a suite of empirical investigations by breaking down error types for more detailed evaluation. To compare two systems, we analyze results on ENG-L2 and JPN-L2 given that they reflect significant advantages of the syntax-based systems over the neural syntax-agnostic system. Note that the syntax-based system here refers to the neural-parser-based one. Finally, a concrete study on the instances in the output is conducted, as to validate conclusions in the previous step.

\begin{table}[H]
\centering
\renewcommand{\arraystretch}{1.1}
\scalebox{0.75}{
\begin{tabular}{m{2cm}m{6.5cm}}
\hline
Operation & Description \\\hline
\begin{tabular}[c]{@{}l@{}}Fix Labels \\ (Fix)\end{tabular} & Correct the span label if its boundary matches gold. \\\hline
\begin{tabular}[c]{@{}l@{}}Move Arg. \\ (Move)\end{tabular} & Move a unique core argument to its correct position. \\\hline
\begin{tabular}[c]{@{}l@{}}Merge Spans \\ (Merge)\end{tabular} & Combine two predicated spans into a gold span if they are separated by at most one word. \\\hline
\begin{tabular}[c]{@{}l@{}}Split Spans \\ (Split)\end{tabular} & Split a predicated span into two gold spans that are separated by at most one word. \\\hline
\begin{tabular}[c]{@{}l@{}}Fix Boundary \\ (Boundary)\end{tabular} & Correct the boundary of a span id its label matches an overlapping gold span. \\\hline
\begin{tabular}[c]{@{}l@{}}Drop Arg. \\ (Drop)\end{tabular} & Drop a predicated argument that does not overlap with any gold span. \\\hline
\begin{tabular}[c]{@{}l@{}}Add Arg. \\ (Add)\end{tabular} & Add a gold argument that does not overlap with any predicated span.\\\hline
\end{tabular}}
\caption{Oracle transformations paired with the relative error reduction after each operation. The operations are permitted only if they do not cause any overlapping arguments}
\label{tb:oracle trans}
\end{table}

\subsubsection{Breaking down Error Types}

\begin{figure*}[]
    \centering
    \begin{subfigure}[t]{0.45\textwidth}
        \centering
        \includegraphics[height=2in]{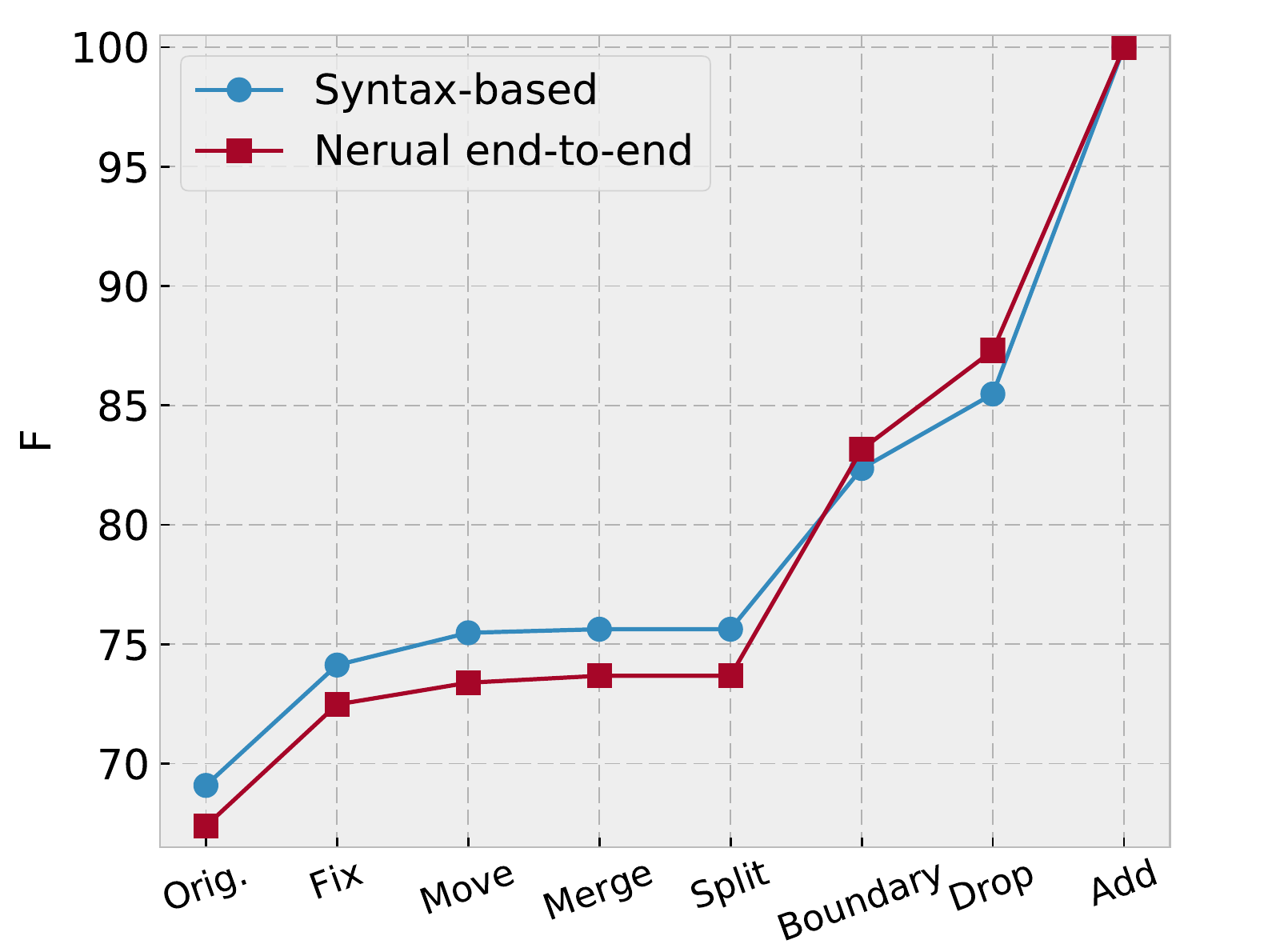}
        \caption{ENG-L2}
    \end{subfigure}%
    ~~~~~~
    \begin{subfigure}[t]{0.45\textwidth}
        \centering
        \includegraphics[height=2in]{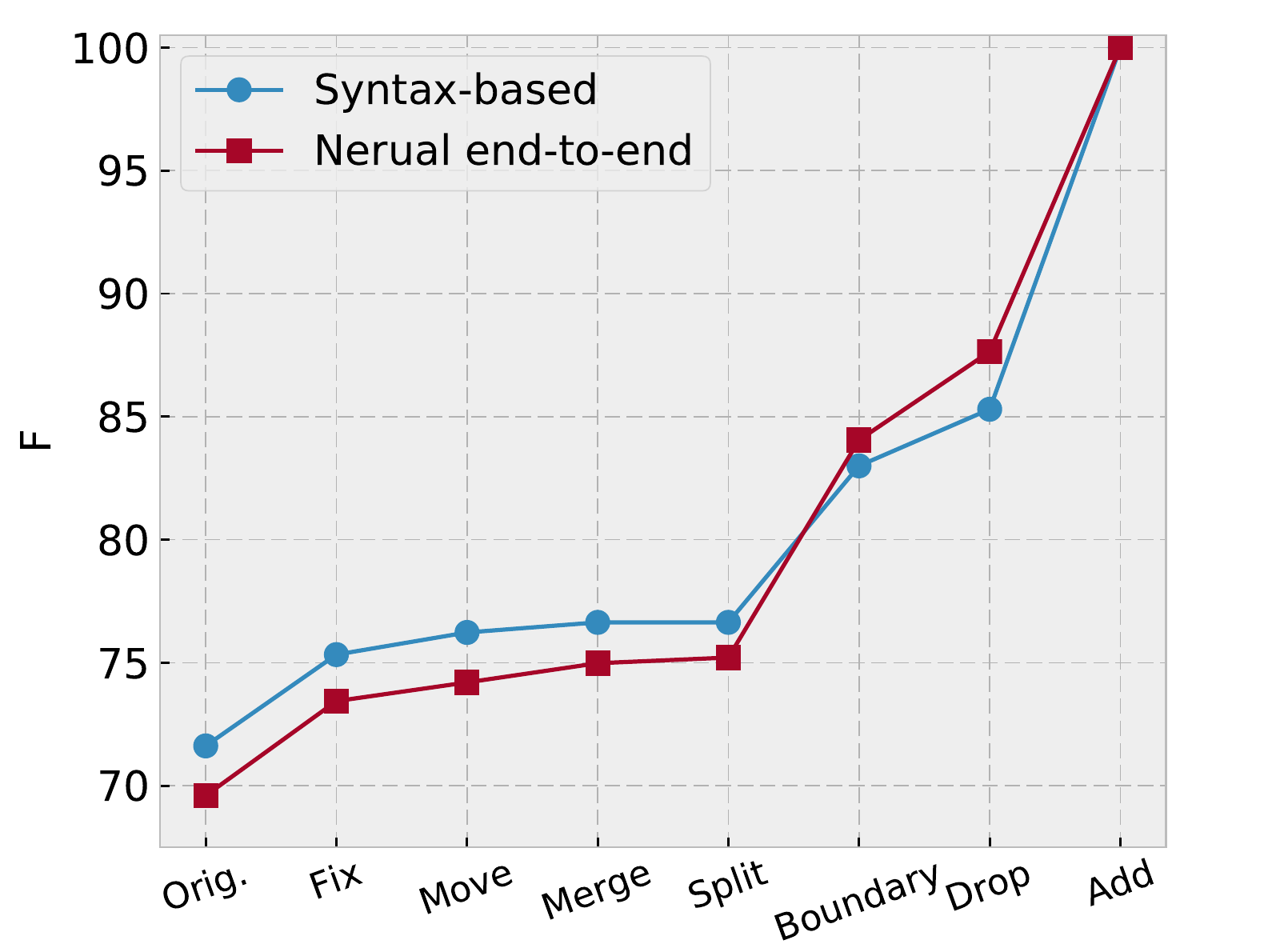}
        \caption{JPN-L2}
    \end{subfigure}
    \caption{Relative improvements of performance after doing each type of oracle transformation in sequence over ENG-L2 and JPN-L2}
    \label{fig:perform l2}
\end{figure*}

We employ 6 oracle transformations designed by \newcite{he2017deep} to fix various prediction errors sequentially (see details in Table \ref{tb:oracle trans}), and observe the relative improvements after each operation, as to obtain fine-grained error types.
Figure \ref{fig:perform l2} compares two systems in terms of different mistakes on ENG-L2 and JPN-L2 respectively. 
After fixing the boundaries of spans, the neural syntax-agnostic system catches up with the other, 
illustrating that though both systems handle boundary detection poorly on the L2 sentences, the neural syntax-agnostic one suffers more from this type of errors.

\begin{figure}[H]
    \centering
    \begin{subfigure}[t]{0.23\textwidth}
        \centering
        \includegraphics[height=1in]{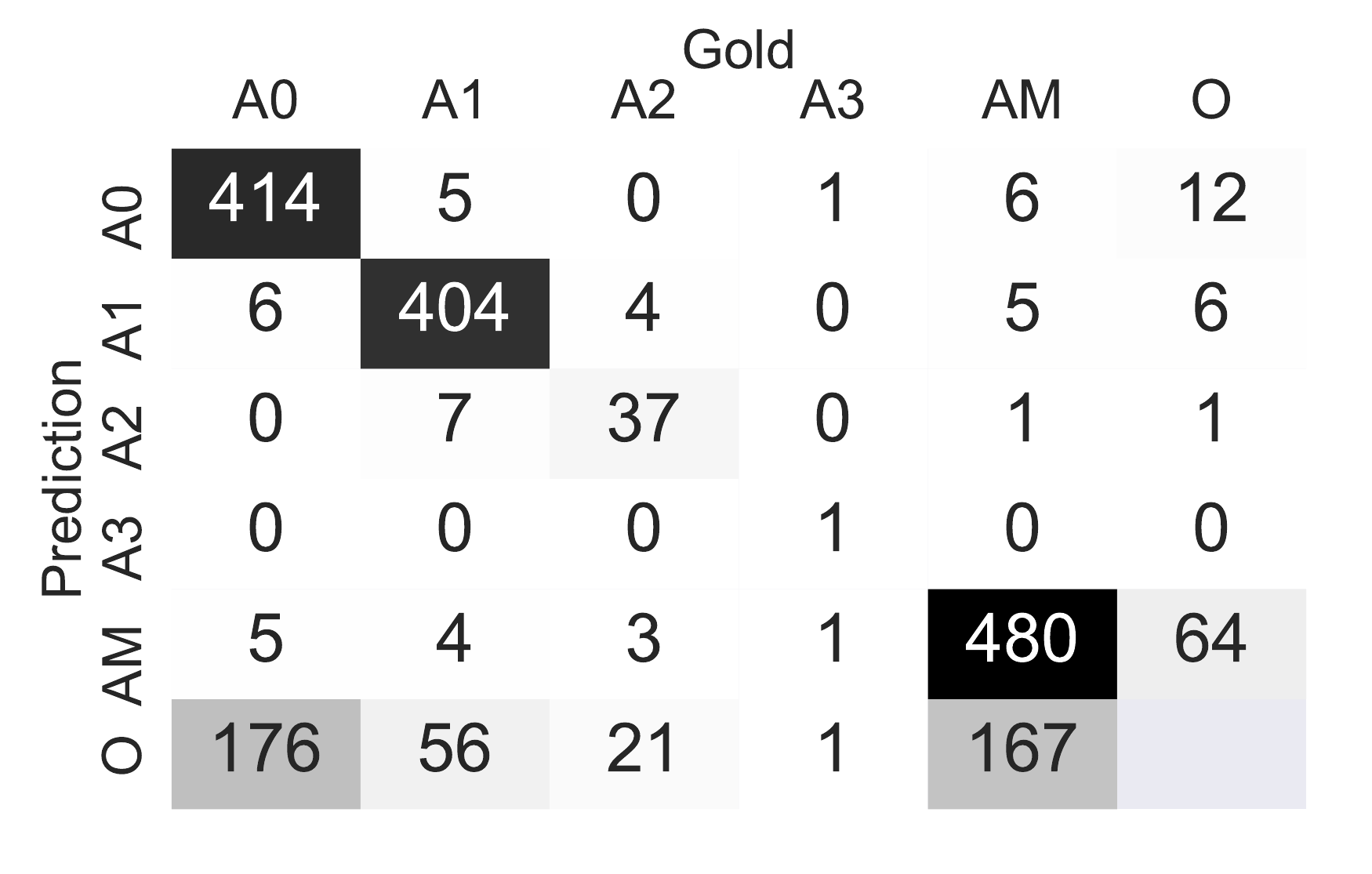}
        \caption{Syntax-based system, L1}
    \end{subfigure}%
    ~ 
    \begin{subfigure}[t]{0.23\textwidth}
        \centering
        \includegraphics[height=1in]{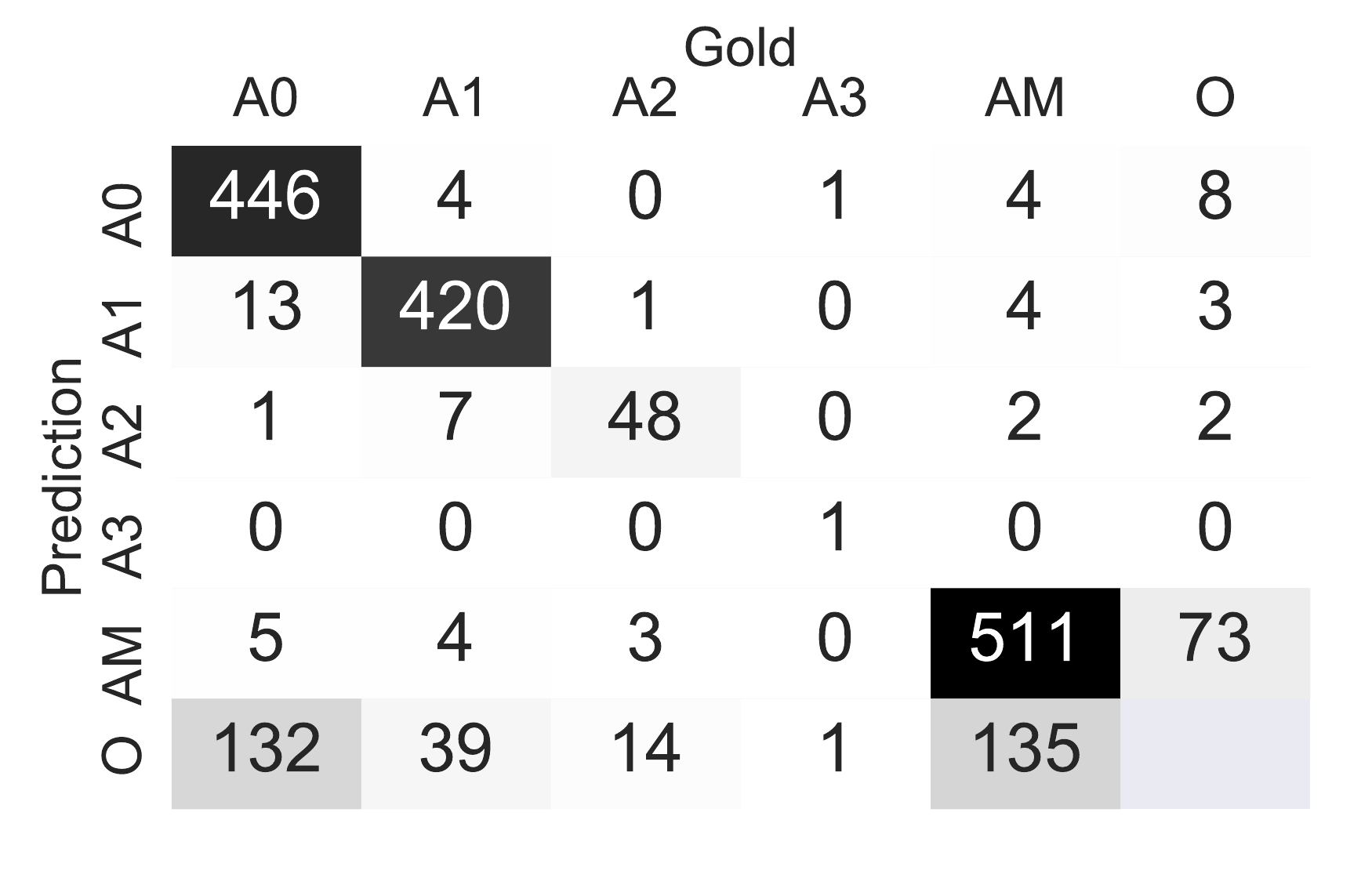}
        \caption{Neural system, L1}
    \end{subfigure}
    \\
    \begin{subfigure}[t]{0.23\textwidth}
        \centering
        \includegraphics[height=1in]{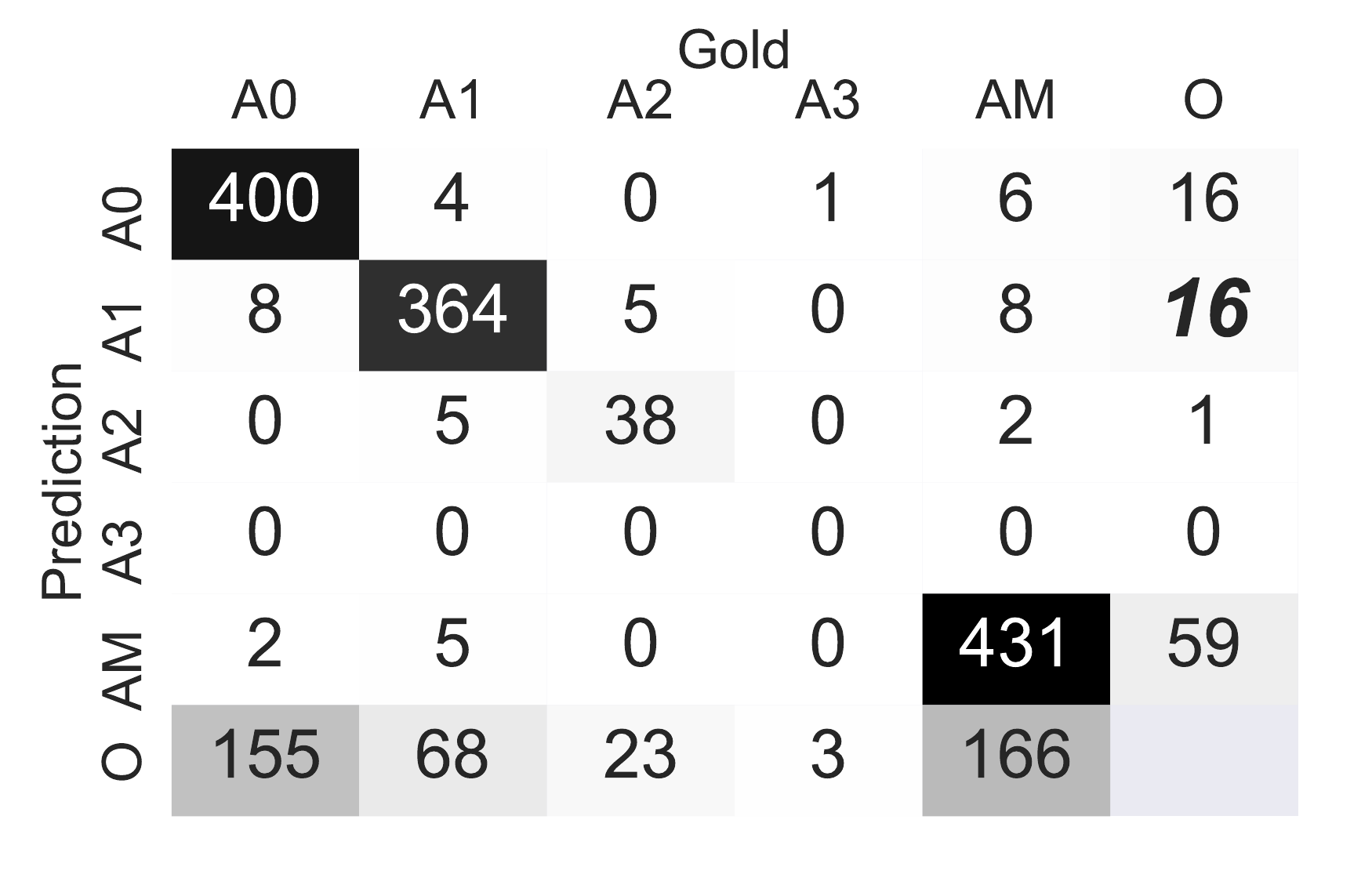}
        \caption{Syntax-based system, L2}
    \end{subfigure}%
    ~ 
    \begin{subfigure}[t]{0.23\textwidth}
        \centering
        \includegraphics[height=1in]{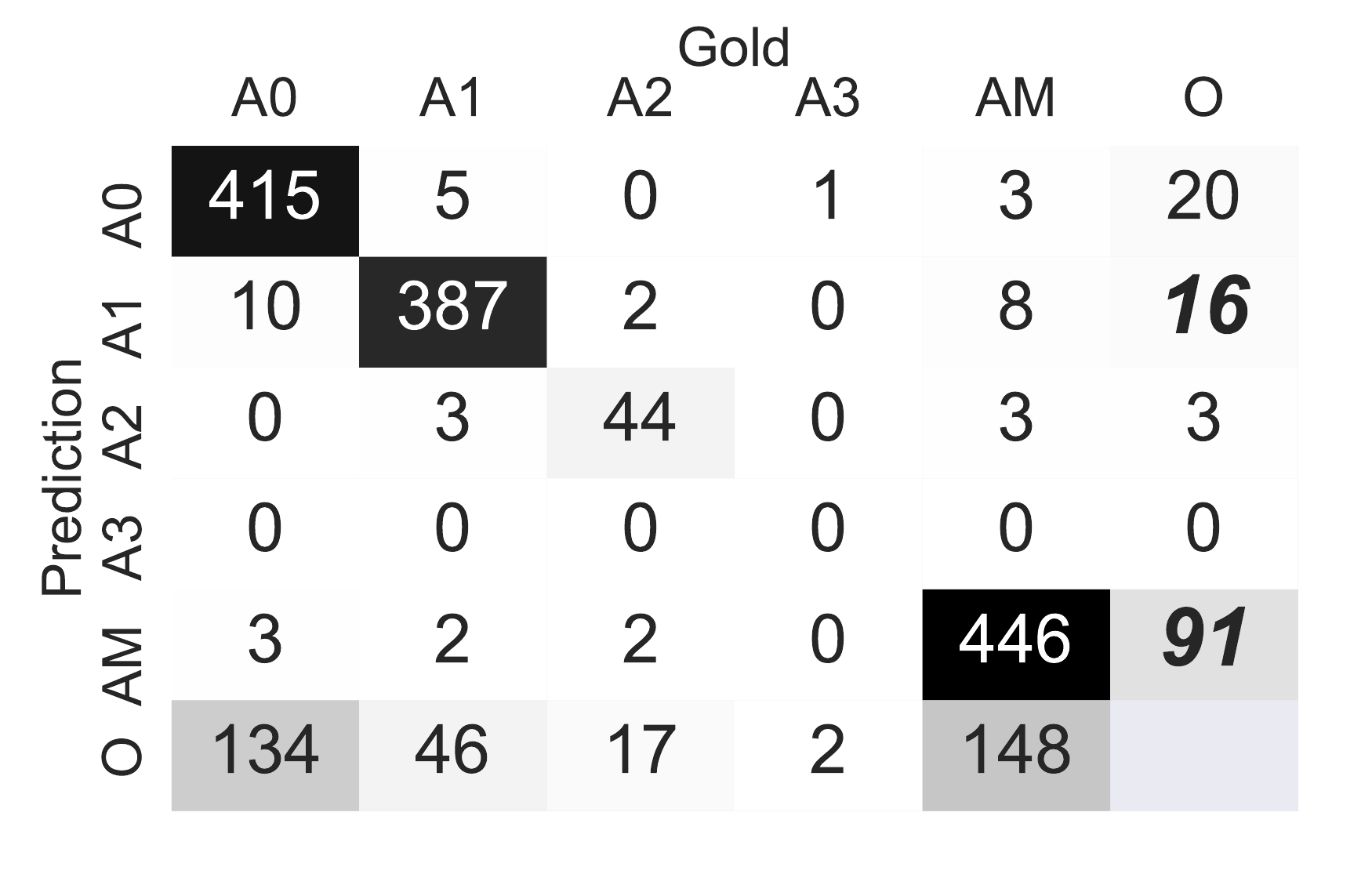}
        \caption{Neural system, L2}
    \end{subfigure}
    \caption{Confusion matrix for each semantic role (here we add up matrices of ENG-L2 and JPN-L2). 
    The predicted labels are only counted in three cases: (1) The predicated boundaries match the gold span boundaries. 
    (2) The predicated argument does not overlap with any the gold span (Gold labeled as ``O''). 
    (3) The gold argument does not overlap with any predicted span (Prediction labeled as ``O'').}
    \label{fig:confusion matrix}
\end{figure}

Excluding boundary errors (after moving, merging, splitting spans and fixing boundaries), we also compare two systems on L2 in terms of detailed label identification, so as to observe which semantic role is more likely to be incorrectly labeled. Figure \ref{fig:confusion matrix} shows the confusion matrices. Comparing (a) with (c) and (b) with (d), we can see that the syntax-based and the neural system often overly label \textit{A1} when processing learner texts. Besides, the neural syntax-agnostic system predicts the adjunct \textit{AM} more than necessary on L2 sentences by 54.24\% compared with the syntax-based one.

\subsubsection{Examples for Validation}

\begin{figure*}[]
	\centering
	\begin{subfigure}[t]{0.45\textwidth}
    	\centering
        \includegraphics[height=1.4in]{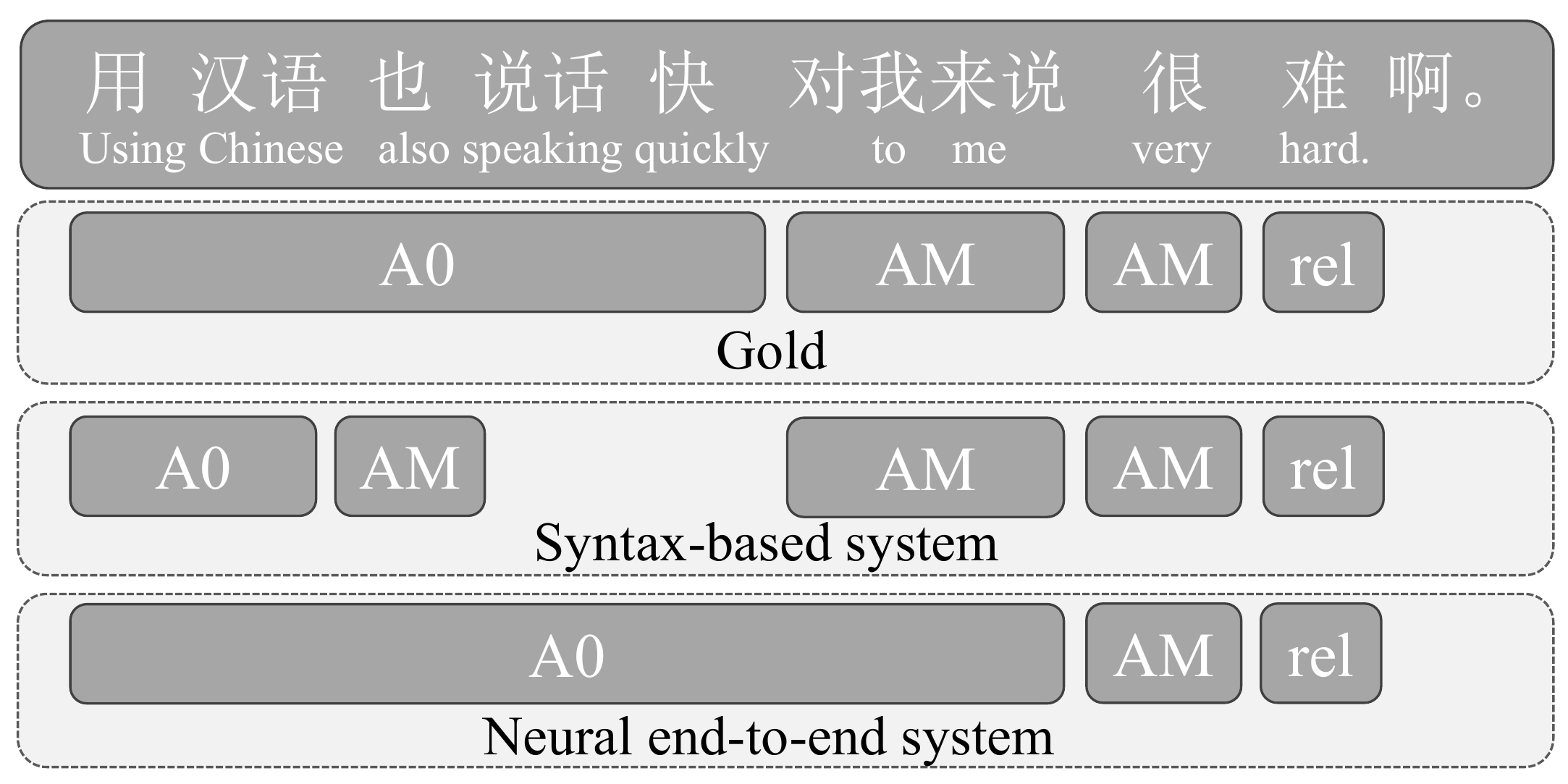}
        \caption{SRL output of both systems for a L2 sentence, ``用汉语也说话快对我来说很难'' (using Chinese and also speaking quickly is very hard for me).}
        \label{fig:output example1}
    \end{subfigure}%
    \hspace{1em}
    \begin{subfigure}[t]{0.45\textwidth}
    	\centering
        \includegraphics[height=1.4in]{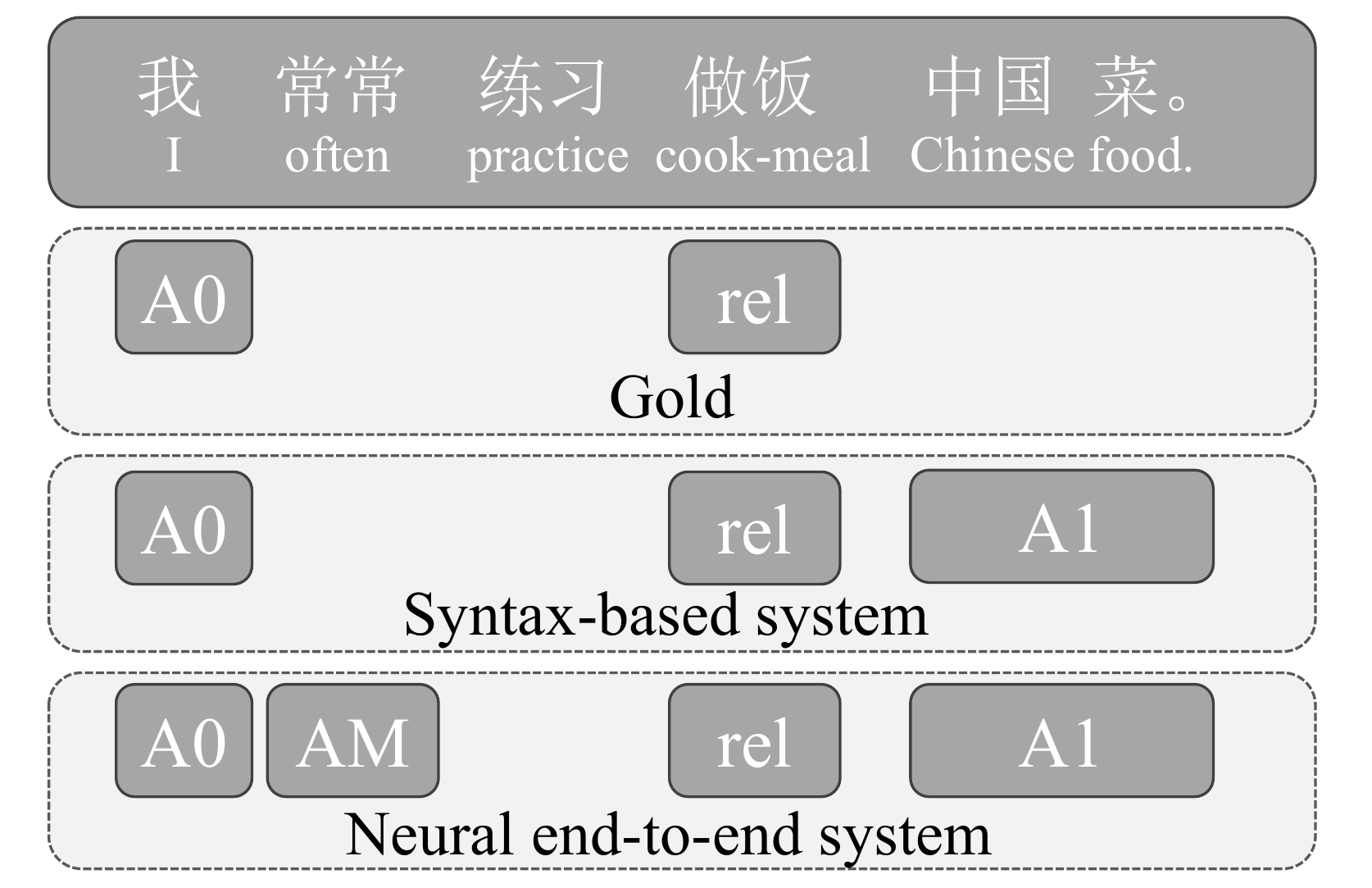}
        \caption{SRL of both systems for a L2 sentence, ``我常常做饭中国菜'' (I often cook Chinese food).}
        \label{fig:output example2}
    \end{subfigure}%
    \\
    \begin{subfigure}[t]{0.45\textwidth}
        \centering
        \begin{spacing}{0.5}
		\scalebox{0.75}{
		\begin{forest}
			[CP [IP [IP, name = [用汉语\\using\\Chinese, roof]] [VP [ADVP, name = [也\\also]] [VP [VP, name = [说话快\\speaking\\quickly,roof]] [VP [PP, name = [对我来说\\for me,roof]] [ADVP , name = [很\\very]] [VP, name = [难\\hard]]]]]] [SP, name = sp[啊\\{\tt MOD}]] [PU, name = pu[。]]]
		\end{forest}}
		\end{spacing}
        \vspace{0.2cm}
        \caption{Syntactic analysis for the sentence in Figure 3a}
        \label{fig:syn analysis1}
    \end{subfigure}
    \hspace{1em}
    \begin{subfigure}[t]{0.45\textwidth}
    	\centering
        \begin{spacing}{0.5}
		\scalebox{0.75}{
		\begin{forest}
			[IP [NP [PN, name = pn2[我\\I]]] [VP [ADVP, name = ad[常常\\often]] [VP [VV, name = vv2[练习\\practice]] [IP [VP [VV, name = vv3[做饭\\cook-meal]] [NP, name = [中国菜\\chinese food,roof]]]]]] [PU, name = pu2[。]]]
		\end{forest}}
		\end{spacing}
        \vspace{0.2cm}
        \caption{Syntactic analysis for the sentence in Figure 3b}
        \label{fig:syn analysis2}
    \end{subfigure}
    \caption{Two examples for SRL outputs of both systems and the corresponding syntactic analysis for the L2 sentences}
\end{figure*}

On the basis of typical error types found in the previous stage, specifically, boundary detection and incorrect labels, we further conduct an on-the-spot investigation on the output sentences.

\paragraph{Boundary Detection} Previous work has proposed that the drop in performance of SRL systems mainly occurs in identifying argument boundaries \cite{marquez2008semantic}. According to our results, this problem will be exacerbated when it comes to L2 sentences, while syntactic structure sometimes helps to address this problem.

Figure \ref{fig:output example1} is an example of an output sentence. 
The Chinese word ``也'' (also) usually serves as an adjunct but is now used for linking the parallel structure ``用\ 汉语\ 也\ 说话\ 快'' (using Chinese also speaking quickly) in this sentence, which is ill-formed to native speakers and negatively affects the boundary detection of \textit{A0} for both systems. 

On the other hand, the neural system incorrectly takes the whole part before ``很\ 难'' (very hard) as \textit{A0}, regardless of the adjunct ``对\ 我\ 来说'' (for me), while this can be figured out by exploiting syntactic analysis, as illustrated in Figure \ref{fig:syn analysis1}. The constituent ``对\ 我\ 来说'' (for me) has been recognized as a prepositional phrase (PP) attached to the VP, thus labeled as \textit{AM}. This shows that by providing information of some well-formed sub-trees associated with correct semantic roles, the syntactic system can perform better than the neural one on SRL for learner texts.

\paragraph{Mistaken Labels}
A second common source of errors is wrong labels, especially for \textit{A1}. Based on our quantitative analysis, as reported in Table \ref{tb:cause of error}, these phenomena are mainly caused by mistakes of verb subcategorization, where the systems label more arguments than allowed by the predicates. Besides, the deep end-to-end system is also likely to incorrectly attach adjuncts \textit{AM} to the predicates.

\begin{table}[h]
\centering
\scalebox{0.9}{
\begin{tabular}{lcc}
\hline
               & \multicolumn{2}{c}{Syntax} \\
               \cline{2-3}
  Cause of error &  YES & NO \\
\hline
Verb subcategorization & 62.50\% & 62.50\% \\
Labeling \textit{A1} to punctuation & 12.50\% & 6.25\% \\
Word order error & 6.25\% & 0.00\% \\
Other types of error & 18.75\% & 31.25\% \\ \hline
\end{tabular}}
\caption{Causes of labeling unnecessary \textit{A1}}
\label{tb:cause of error}
\end{table}

Figure \ref{fig:output example2} is another example. The Chinese verb ``做饭'' (cook-meal) is intransitive while this sentence takes it as a transitive verb, which is very common in L2. Lacking in proper verb subcategorization, both two systems fail to recognize those verbs allowing only one argument and label the \textit{A1} incorrectly.

As for \textit{AM}, the neural system mistakenly adds the adjunct to the predicate, which can be avoided by syntactic information of the sentence shown in Figure \ref{fig:syn analysis2}. The constituent ``常常'' (often) are adjuncts attached to VP structure governed by the verb ``练习''(practice), which will not be labeled as \textit{AM} in terms of the verb ``做饭''(cook-meal). In other words, the hierarchical structure can help in argument identification and assignment by exploiting local information.

\section{Enhancing SRL with L2-L1 Parallel Data}

We explore the valuable information about the semantic coherency encoded in the L2-L1 parallel data to improve SRL for learner Chinese.
In particular, we introduce an agreement-based model to search for high-quality automatic syntactic and semantic role annotations, and then use these annotations to retrain the two parser-based SRL systems.

\subsection{The Method}

For the purpose of harvesting the good automatic syntactic and semantic analysis, 
we consider the consistency between the automatically produced analysis of 
a learner sentence and its corresponding {\it well-formed} sentence.
Determining the measurement metric for comparing predicate--argument structures, however, presents another challenge, 
because the words of the L2 sentence and its L1 counterpart do not necessarily match. 
To solve the problem, we use an automatic word aligner.
BerkeleyAligner\footnote{\url{code.google.com/archive/p/berkeleyaligner/}} \citep{Liang-Taskar-Klein:2006:Alignment}, a state-of-the-art tool for obtaining a word alignment, is utilized.

The metric for comparing SRL results of two sentences is based on recall of $\langle w_p,w_a,r\rangle$ tuples,
where $w_p$ is a predicate, $w_a$ is a word that is in the argument or adjunct of $w_p$  and $r$ is the corresponding role.
Based on a word alignment, we define the shared tuple as a mutual tuple between two SRL results of an L2-L1 sentence pair, meaning that both the predicate and argument words 
are aligned respectively, and their role relations are the same.
We then have two recall values:
\begin{itemize}
\item L2-recall is (\# of shared tuples) / (\# of tuples of the result in L2)
\item L1-recall is (\# of shared tuples) / (\# of tuples of the result in L1)
\end{itemize}

In accordance with the above evaluation method, 
we select the automatic analysis of highest scoring sentences and use them to expand the training data.
Sentences whose L1 and L2 recall are both greater than a threshold $p$ are taken as good ones.
A parser-based SRL system consists of two essential modules: a syntactic parser and a semantic classifier.
To enhance the syntactic parser, the automatically generated syntactic trees of the sentence pairs that exhibit high semantic consistency are directly used to extend training data. 
To improve a semantic classifier, besides the consistent semantic analysis, we also use the outputs of the L1 but not L2 data which are generated by the neural syntax-agnostic SRL system.


\subsection{Experimental Setup}

Our SRL corpus contains 1200 sentences in total that can be used as an evaluation for SRL systems.
We separate them into three data sets.
The first data set is used as development data, which contains 50 L2-L1 sentence pairs for each language and 200 pairs in total.
Hyperparameters are tuned using the development set.
The second data set contains all other 400 L2 sentences, which is used as test data for L2.
Similarly, all other 400 L1 sentences are used as test data for L1.

\begin{table}[]
\centering
\scalebox{0.99}{
\begin{tabular}{l|cc}
\hline
       & ENG & JPN \\
       \hline
\#{All sentence pairs} & 310,075 & 484,140 \\
\#{Selected} ($p=0.9$) & 36,979 & 41,281 \\
\hline
\end{tabular}}
\caption{Statistics of unlabeled data.}
\label{tb:pool}
\end{table}

The sentence pool for extracting retraining annotations includes all English- and Japanese-native speakers' data along with its corrections.
Table \ref{tb:pool} presents the basic statistics. Around 8.5 -- 11.9\% of the sentence can be taken as high L1/L2 recall sentences, which serves as a reflection that argument structure is vital for language acquisition and difficult for learners to master, as proposed in \newcite{vazquez2004learning} and \newcite{shin2010contribution}.
The threshold ($p=0.9$) for selecting sentences is set upon the development data.
For example, we use additional 156,520 sentences to enhance the Berkeley parser.

\subsection{Main Results}

Table \ref{tb:retrained} summarizes the SRL results of the baseline PCFGLA-parser-based model as 
well as its corresponding retrained models.
Since both the syntactic parser and the SRL classifier can be retrained and thus enhanced, we report the individual impact as well as the combined one.
We can clearly see that when the PCFGLA parser is retrained with the SRL-consistent sentence pairs, it is able to provide better SRL-oriented syntactic analysis for the L2 sentences as well as 
their corrections, which are essentially L1 sentences.
The outputs of the L1 sentences that are generated by the deep SRL system are also useful for improving the linear SRL classifier.
A non-obvious fact is that such a retrained model yields better analysis for not only L1  but also L2 sentences.
Fortunately, combining both results in further improvement.

\begin{table}[H]
\centering
\scalebox{0.9}{
\begin{tabular}{llccc}
\hline
 &             & P & R & F \\
\hline
\multirow{4}{*}{L2} & Baseline & 76.50 &  62.86 &  69.01\\
&Parser-retrained & 78.74   & 65.87   & 71.73 \\
&Classifier-retrained & 77.11   & 64.17   & 70.05\\
&Both retrained & 78.72 &  66.43  & 72.06\\
\hline
\multirow{4}{*}{L1}  & Baseline & 80.70 &  66.36 &  72.83 \\
&Parser-retrained & 82.20  &  68.26  & 74.59 \\
&Classifier-retrained & 80.98  &  67.57  & 73.67 \\
&Both retrained &  82.04  & 68.79  & 74.83 \\
\hline
\end{tabular}}
\caption{Accuracies different PCFGLA-parser-based models on the two test data sets.}
\label{tb:retrained}
\end{table}

Table \ref{tb:retrained-neural} shows the results of the parallel experiments based on the neural parser.
Different from the PCFGLA model, the SRL-consistent trees only yield a slight improvement on the L2 data.
On the contrary, retraining the SRL classifier is much more effective.
This experiment highlights the different strengths of different frameworks for parsing.
Though for {\it standard in-domain} test, the neural parser performs better and thus is more and more popular, for 
some other scenarios, the PCFGLA model is stronger.

\begin{table}[H]
\centering
\scalebox{0.9}{
\begin{tabular}{llccc}
\hline
 &             & P & R & F \\
\hline
\multirow{4}{*}{L2} & Baseline & 78.26 & 63.38 & 70.04  \\
&Parser-retrained & 78.19 & 63.78 & 70.25 \\
&Classifier-retrained & 78.88 & 64.65 & 71.06 \\
&Both retrained &  78.61 & 65.34 & 71.36 \\
\hline
\multirow{4}{*}{L1}  & Baseline & 82.17 & 66.80 & 73.69 \\
&Parser-retrained & 81.95 & 66.92 & 73.68 \\
&Classifier-retrained & 82.08 & 67.69 & 74.20 \\
&Both retrained &  82.20 & 67.85 & 74.34 \\
\hline
\end{tabular}}
\caption{Accuracies of different neural-parser-based models on the two test data sets.}
\label{tb:retrained-neural}
\end{table}

Table \ref{tb:f-score role types} further shows F-scores for the baseline and the both-retrained model relative to each role type in detail. Given that the F-scores for both models are equal to 0 on \textit{A3} and \textit{A4}, we just omit this part. From the figure we can observe that, all the semantic roles achieve significant improvements in performances.

\begin{table}[H]
\centering
\scalebox{0.85}{
\begin{tabular}{llllllll}
\hline
 &  & \multicolumn{1}{c}{\textit{A0}} & \multicolumn{1}{c}{\textit{A1}} & \multicolumn{1}{c}{\textit{A2}} &  \multicolumn{1}{c}{\textit{AM}} \\ \hline
\multirow{2}{*}{L2} & Baseline & 67.95 & 71.21 & 51.43  & 70.20 \\
 & Both retrained & 70.62 & 74.75 & 64.29  & 72.22 \\ \hline
\multirow{2}{*}{L1} & Baseline & 69.49 & 79.78 & 61.84 & 71.74 \\
 & Both retrained & 73.15 & 80.90 & 63.35  & 73.02\\\hline
\end{tabular}}
\caption{F-scores of the baseline and the both-retrained models relative to role types on the two data sets. We only list results of the PCFGLA-parser-based system.}
\label{tb:f-score role types}
\end{table}

\section{Conclusion}

Statistical models of annotating learner texts are making rapid progress.
Although there have been some initial studies on defining annotation specification as well as corpora for syntactic analysis, there is almost no work on semantic parsing for interlanguages.
This paper discusses this topic, taking Semantic Role Labeling as a case task and learner Chinese as a case language. 
We reveal three unknown facts that are important towards a deeper analysis of learner languages: (1) the robustness of language comprehension for interlanguage,
(2) the weakness of applying L1-sentence-trained systems to process learner texts, 
and (3) the significance of syntactic parsing and L2-L1 parallel data in building more generalizable SRL models that transfer better to L2. 
We have successfully provided a better SRL-oriented syntactic parser as well as a semantic classifier for processing the L2 data by exploring L2-L1 parallel data, supported by a significant numeric improvement over a number of state-of-the-art systems. 
To the best of our knowledge, this is the first work that demonstrates the effectiveness of large-scale L2-L1 parallel data to enhance the NLP system for learner texts.

\section*{Acknowledgement}

This work was supported by the National Natural Science Foundation of China (61772036, 61331011)
and the Key Laboratory of Science, Technology and Standard in Press Industry (Key Laboratory of Intelligent Press Media Technology). We thank the anonymous reviewers and for their helpful comments. We also thank Nianwen Xue for useful comments on the final version.
Weiwei Sun is the corresponding author.

\end{CJK*}
\bibliographystyle{acl_natbib_nourl}
\bibliography{references}

\end{document}